\pdfoutput=1

\documentclass[11pt]{article}

\usepackage[]{naacl2022}

\usepackage{times}
\usepackage{latexsym}

\usepackage[T1]{fontenc}

\usepackage[utf8]{inputenc}

\usepackage{microtype}
\usepackage{array}
\usepackage{multirow}
\usepackage{graphicx}
\usepackage{algorithm}
\usepackage{algpseudocode}
\usepackage{enumitem}

\usepackage{amsmath}
\usepackage{amssymb}
\usepackage{subcaption}

\DeclareMathOperator*{\argmax}{argmax}

\usepackage{graphicx}
\usepackage{dblfloatfix} 
\usepackage{caption}
\usepackage[labelformat=simple]{subcaption}

\newcommand{\revised}[1]{\textcolor{black}{#1}}

\title{MedJEx: A Medical Jargon Extraction Model with Wiki's Hyperlink Span and Contextualized Masked Language Model Score}

\author{\begin{tabular}[c]{@{}c@{}}Sunjae Kwon$^{1}$, Zonghai Yao$^{1}$, Harmon S. Jordan$^{2}$, \\David A. Levy$^{3}$, Brian Corner$^{4}$, Hong Yu$^{1,3,4,5}$\end{tabular} \\
  \begin{tabular}[c]{@{}c@{}} 
  $^{1}$UMass Amherst, $^{2}$Health Research Consultant, \\ $^{3}$UMass Lowell, $^{4}$UMass Medical School, $^{5}$U.S. Department of Veterans Affairs
  \end{tabular}
  \\
  \begin{tabular}[c]{@{}c@{}}
  \texttt{sunjaekwon@umass.edu, zonghaiyao@umass.edu, harmon.s.jordan@gmail.com,}\\\texttt{david\_levy@uml.edu, brian.corner@umassmed.edu, hong\_yu@uml.edu}\end{tabular} \\}

\begin{document}
\maketitle
\begin{abstract}

%
This paper proposes a new natural language processing (NLP) application for identifying medical jargon terms potentially difficult for patients to comprehend from electronic health record (EHR) notes. We first present a novel and publicly available dataset with expert-annotated medical jargon terms from 18K+ EHR note sentences ($MedJ$). Then, we introduce a novel medical jargon extraction ($MedJEx$) model which has been shown to outperform existing state-of-the-art NLP models. First, MedJEx improved the overall performance when it was trained on an auxiliary Wikipedia hyperlink span dataset, where hyperlink spans provide additional Wikipedia articles to explain the spans (or terms), and then fine-tuned on the annotated MedJ data. Secondly, we found that a contextualized masked language model score was beneficial for detecting domain-specific unfamiliar jargon terms. Moreover, our results show that training on the auxiliary Wikipedia hyperlink span datasets improved six out of eight biomedical named entity recognition benchmark datasets. Both MedJ and MedJEx are publicly available \footnote{Code and the pre-trained models are available at \url{https://github.com/MozziTasteBitter/MedJEx}.}.   
\end{abstract}

\section{Introduction} 
\revised{Allowing patients to access their electronic health records (EHRs) represents a new and personalized communication channel that has the potential to improve patient involvement in care and assist communication between physicians, patients, and other healthcare providers \citep{baldry1986giving, schillinger2009effects}. However, studies showed that patients do not understand medical jargon in their EHR notes \citep{chen2018natural}.}

To improve patients' EHR note comprehension, it is important to identify medical jargon terms that are difficult for patients to understand. Unlike the traditional concept identification or named entity recognition (NER) tasks, where the tasks mainly center on semantic salient entities, detecting such medical jargon terms takes into consideration the perspective of user comprehension. Traditional NER approaches such as using comprehensive clinical terminological resources (e.g., the Unified Medical Language System (UMLS) \citep{bodenreider2004unified}) would identify terms such as ``water'' and ``fat'', which are not considered difficult for patients to comprehend. Meanwhile, using term frequency (TF) as the proxy for medical jargon term identification will miss outliers such as "shock," which is a term frequently used in the open domain	with its common sense: ``a sudden upsetting or surprising event or experience." However, EHR notes incorporate its uncommon sense: ``a medical condition caused by severe injury, pain, loss of blood, or fear that slows down the flow of blood.'' \cite{Shock}. Thus, ``shock'' should be identified as a jargon term from EHR notes since it would be difficult for patients to comprehend, even though its TF is high. 
In this study, we propose a natural language processing (NLP) system that can identify such outlier jargon from EHR notes through a novel method for homonym resolution. 

We first expert-annotated de-identified EHR note sentences for medical jargon terms judged to be difficult to comprehend. This resulted in the Medical Jargon Extraction for Improving EHR Text Comprehension (MedJ) dataset, which comprises 18,178 sentences and 95,393 medical jargon terms. We then present a neural network-based medical jargon extraction (MedJEx) model to identify the jargon terms.

To ameliorate the limited training-size issue, we propose a novel transfer learning-based framework \citep{tan2018survey} utilizing auxiliary Wikipedia (Wiki) hyperlink span datasets (WikiHyperlink), where the span terms link to different Wiki articles \cite{mihalcea2007wikify}. Although medical jargon extraction and WikiHyperlink recognition seem to be two different applications, they share similarities. The role of hyperlinks is to help a reader to understand an Wiki article. Thus, "difficult to understand" concepts in the Wiki article may be more likely to have hyperlinks. Therefore, we hypothesize that large-scale hyperlink span information from Wiki can be advantageous for our models of medical jargon extraction. Our results show that models trained on WikiHyperlink span datasets indeed substantially improved the performance of MedJEx. Moreover, we also found that such auxiliary learning improved six out of the eight benchmark datasets of biomedical NER tasks.

\revised{To detect outlier homonymous terms such as ``shock'', we deployed an approach
inspired by masking probing \citep{petroni2019language}, a method for evaluating linguistic knowledge of large-scale pre-trained language models (PLMs). \citet{meister2022analyzing} suggests PLMs are beneficial for predicting the reading time, with longer reading time indicates difficult for indicating difficulty in understanding.  
In our work, we propose a contextualized masked language model (MLM) score feature to tackle the homonym challenge. Note that models will recognize the sense of a word or phrase using contextual information. Since PLMs calculate the probability of masked words in consideration of context, we hypothesize that PLMs trained in the open-domain corpus would predict poorly masked medical jargon if senses are distributed differently between the open domain and clinical domain corpora.}

We conducted experiments on four state-of-the-art PLMs, namely BERT \citep{devlin2019bert}, RoBERTa \citep{liu2019roberta}, BioClinicalBERT \citep{alsentzer2019publicly} and BioBERT \citep{lee2020biobert}. Experimental results show that when both of the methods are combined, the medical jargon extraction performance is improved by 2.44\%p in BERT, 2.42\%p in RoBERTa, 1.56\%p in BioClinicalBERT, and 1.19\%p in BioBERT. 

Our contributions are as follows: 
\begin{itemize}
    \item We \textbf{propose a novel NLP task} for identifying medical jargon terms potentially difficult for patients to comprehend from EHR notes.
    \item We will release \textbf{MedJ}, an expert-curated 18K+ sentence dataset for the MedJEx task.
    \item \revised{We introduce \textbf{MedJEx}, a medical jargon extraction model. Herein, MedJEx was first trained with the auxiliary WikiHyperlink span dataset before being fine-tuned on the MedJ dataset. It uses MLM score feature for homonym resolution.} 
    \item The experimental results show that training on the Wiki’s hyperlink span datasets consistently improved the performance of not only MedJ but also six out of eight BioNER benchmarks. In addition, our qualitative analyses show that the MLM score can complement the TF score for detecting the outlier jargon terms.  
    
\end{itemize}

\begin{figure*}[!ht]
\centering
\includegraphics[width=\linewidth]{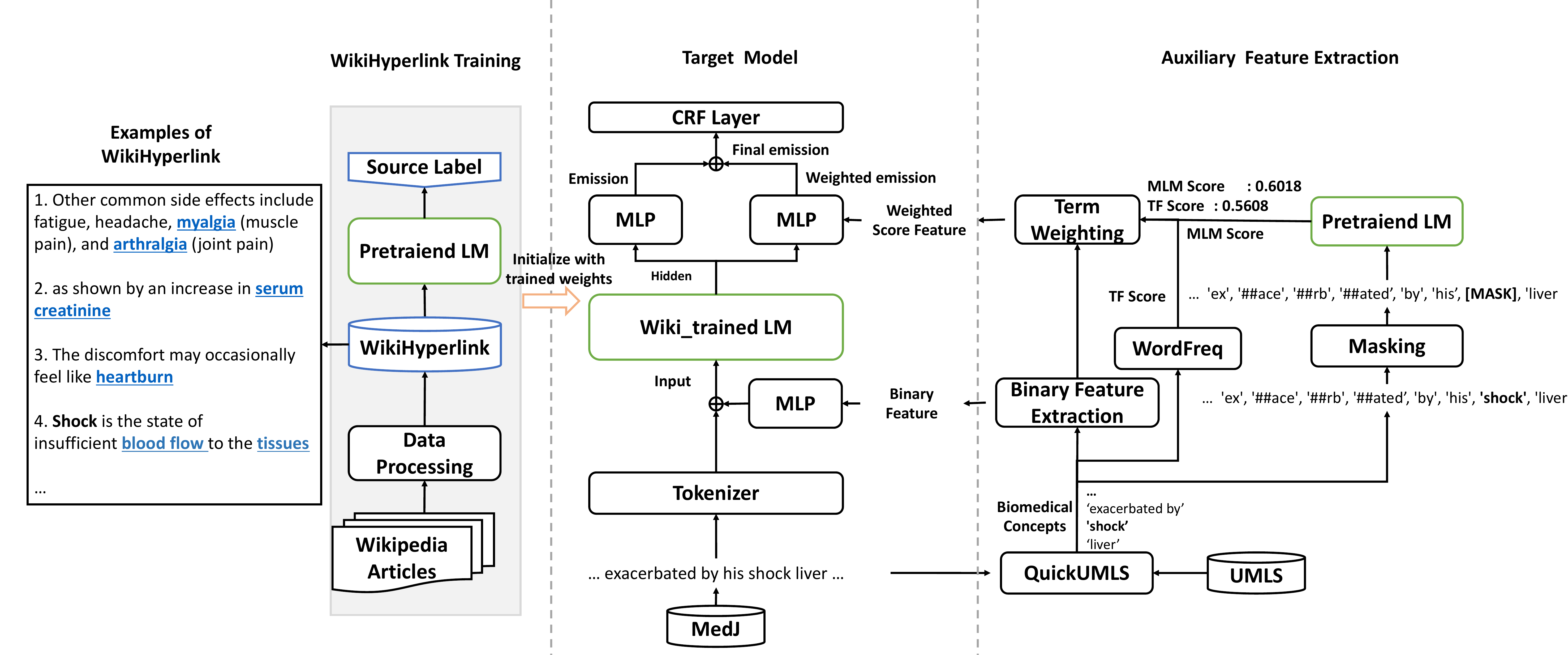}
\caption{This figure demonstrates the overall architecture of MedJEx. There are three components in MedJEx: 1) WikiHyperlink training, 2) auxiliary feature extraction and 3) target model. First, in WikiHyperlink training, we extract hyperlink spans from Wikipedia articles. The examples shows that hyperlink spans (blue colored) represent medical jargons, and ignore easier medical terms such as ``fatigue'' and ``headache''. Then, the pretrained language model (LM) is trained with WikiHyperlink. In auxiliary feature extraction, we can see that MLM score of medical jargon ``shock'' shows relatively high TF and MLM scores, indicating that the MLM score can help detect the medical jargon. Finally, the weight parameters of Wiki-trained LM in the target model are initialized with trained parameters of pretrained LM of WikiHyperlink training. Then, the model is finetuned with MedJ.}

\label{fig:overall_structure}
\end{figure*}

\section{Related Work} 
\revised{In principle, MedJEx is related to text simplification \citep{kandula2010semantic}. None of the previous work \citep{abrahamsson2014medical, qenam2017text, nassar2019neural} identified terms that important for comprehension.} %

On the other hand, MedJEx is relevant to BioNER, a task for identifying biomedical named entities such as $disease$, $drug$, and $symptom$ from medical text. There are several benchmark corpora, including i2b2 2010 \citep{patrick2010high}, ShARe/CLEF 2013 \cite{zuccon2013identify}, and MADE \cite{jagannatha2019overview}, all of which were developed solely based on clinical importance. In contrast, $MedJ$ is patient-centered, taking into consideration of patients' comprehension. Identifying BioNER from medical documents has been an active area of research. Earlier work such as the MetaMap \cite{aronson2001effective}, used linguistic patterns, either manually constructed or learned semi-automatically, to map free text to external knowledge resources such as UMLS \cite{lindberg1993unified}.
The benchmark corpora have promoted supervised machine learning approaches including conditional random fields and deep learning approaches \cite{jagannatha2019overview}.

Key phrase extraction in the medical domain is another related task. It identifies important phrases or clauses that represent topics \citep{hulth2003improved}. In previous studies, key phrases were extracted using features such as TF, word stickiness, and word centrality \citep{saputra2018keyphrases}. \citet{chen2017unsupervised} proposed an unsupervised learning based method to elicit important medical terms from EHR notes using MetaMap \citep{demner2017metamap} and various weighting features such as TextRank \citep{mihalcea2004textrank} and term familiarity score \citep{zeng2007making}. In another work, \citet{chen2017ranking} proposed an adaptive distant supervision based medical term extraction approach that utilizes consumer health vocabulary \citep{zeng2006exploring} and a heuristic rule to distantly label medical jargon training datasets. A key phrase extraction method using a large-scale pretrained model is being actively studied \citep{soundarajan2021sparclink}.

Unlike the previous BioNER or key phrase identification applications, identifying medical jargon terms is important for patients' comprehension of their EHR notes and represents a novel NLP application. However, not all medical entities are unfamiliar to patients. The brute force approach of capturing every medical entity, the approaches of existing BioNER and key phrase identification applications, may bring about confusion to patients. On the other hand, undetected medical jargon terms will reduce patients' EHR note comprehension. In this paper, we propose MedJEx, a novel application that identifies medical jargon terms important for patients' comprehension. \revised{Once jargon terms are identified, interventions such as linking the jargon terms to lay definitions can help improve comprehension.}

\section{Dataset Construction}

This work has two different datasets: 1) \textit{MedJ} for medical jargon extraction and 2) \textit{Wiki's hyperlink span (WikiHyperlink)} dataset for transfer learning. 

\subsection{MedJ}
\subsubsection{Data Collection} \revised{The source of the dataset is a collection of publicly available deidentified EHR notes from hospitals affiliated with the University of Pittsburg Medical Center. Herein, 18,178 sentences were randomly sampled and domain-experts then annotated the sentences for medical jargon \footnote{Using these data requires a license agreement.}.}

\subsubsection{Data Annotation} Domain-experts read each sentence and identified as medical jargon terms that would be considered difficult to comprehend for anyone no greater than a 7th grade education\footnote{The rule of thumb is that if a candidate term has a lay definition comprehensible to a 4-7th grader as judged by Flesch-Kincaid Grade Level \citep{solnyshkina2017evaluating}, the candidate term is included as a jargon term.}. Overall, 96,479 medical jargon terms have been annotated by complying with the following annotation guideline.

\paragraph{Annotation Guideline}
\revised{The dataset was annotated for medical jargon by six domain experts from medicine, nursing, biostatistics, biochemistry, and biomedical literature curation \footnote{The annotator agreement scores can be found in Appendix~\ref{apx:annotation_reliability}.}. Herein, the annotators applied the following rules for identifying what was jargon:}

\noindent \textbf{Rule 1.} Medical terms that would \textbf{not be recognized by about 4 to 7th graders}, or that \textbf{have a different meaning in the medical context than in the lay context (homonym)} were defined. For example:

\begin{itemize}
    \item accommodate: When the eye changes focus from far to near.
    \item antagonize: A drug or substance that stops the action or effect of another substance.
    \item resident: A doctor who has finished medical school and is receiving more training.
    \item formed: Stool that is solid.
\end{itemize}

\noindent\textbf{Rule 2.} Terms that are not strictly medical, but are \textbf{frequently used in medicine} were defined. For example:

\begin{itemize}
    \item ``aberrant'', ``acute'', ``ammonia'', ``tender'', ``intact'', ``negative'', ``evidence''
\end{itemize}
 
\noindent\textbf{Rule 3.} When jargon words are \textbf{commonly used together, or together they mean something distinct or are difficult to quickly understand from the individual parts}, they were defined. For example:

\begin{itemize}
    \item vascular surgery: Medical specialty that performs surgery on blood vessels.
    \item airway protection: Inserting a tube into the windpipe to keep it wide open and prevent vomit or other material from getting into the lungs.
    \item posterior capsule: The thin layer of tissue behind the lens of the eye. It can become cloudy and blur vision.
    \item right heart: The side of the heart that pumps blood from the body into the lungs.
    \item intracerebral hemorrhage: A stroke.
\end{itemize}

\noindent\textbf{Rule 4.} Terms whose \textbf{definitions are widely known} (e.g., by a 3rd grader) do NOT need to be defined. For example:

\begin{itemize}
    \item “muscle”, “heart”, “pain”, “rib”, “hospital”
\end{itemize}

\textbf{Rule 4.1} When in doubt, define the term. For example: 
\begin{itemize}
    \item “colon”, “immune system”
\end{itemize}

\subsubsection{Data Cleaning} 
First, we cleaned up overlapped (tumor suppressor \textit{gene}, \textit{gene} deletion) or nested (\textit{vitamin D}, 25-hydroxy \textit{vitamin D}) jargon. We chose the longest jargon terms among nested or overlapped jargon terms. For example, we chose "tumor suppressor gene" as a jargon term, not its nested term "tumor." In all, MedJ contains a total of 95,393 context-dependent jargon terms which we used as the gold standard for training and evaluation of the MedJEx model. The 95,393 jargon terms represent a total of 12,383 unique jargon terms. 

\subsection{WikiHyperlink}
From a Wiki dump data\footnote{\url{https://dumps.wikimedia.org/enwiki/20211001/}}, we first cleaned and elicited text by using Wikiextractor \citep{Wikiextractor2015}. Then, we extracted hyperlink spans with the BeautifulSoup \citep{richardson2007beautiful} module. Wiki articles were split into sentences with the Natural Language Toolkit \citep{bird2009natural}, then the sentences were split into tokens with the PLM tokenizer. \revised{Overall, WikiHyperlink contains more than 114M sentences, 13B words, and 99M hyperlink spans.} Finally, the source data consists of the sequence input of the token and hyperlink labels represented in the standard BIOES format \citep{yang2018design}.

\section{MedJEx Model} 

Figure~\ref{fig:overall_structure} is an overview of MedJEx. First, we trained PLMs with WikiHyperlink (Wiki-trained). Then, the Wiki-trained model was transferred to the target model that we propose by initializing the target model with the weight parameters of the Wiki-trained model. Finally, we fine-tuned the target model with our expert-annotated dataset. \revised{Note that, since the pretrain corpora of PLMs used in this work include the Wiki corpus, we noticed that the performance change should derive from the added labels (hyperlink spans).} Herein, we extracted UMLS concepts and used them as auxiliary features.

\subsection{Wiki's Hyperlink Span Prediction for Transfer Learning Framework}
\revised{Although MedJ is a high-quality and a large scale expert-labeled dataset, deep learning models could improve performance with additional data. However, annotation is very expensive. Transfer learning is one of the effective ways to mitigate the challenge \citep{ruder2019neural, mao2020survey}.} In this paper, we propose to utilize Wiki's articles and hyperlink span as source data. We assumed that hyperlink spans are similar to medical jargon: readers need to read the hyperlinked articles to understand the span concepts \citep{mihalcea2007wikify}. Indeed, in the example sentence in Figure~\ref{fig:overall_structure}, we can see that some difficult biomedical concepts are hyperlinked, but easier concepts such as "fatigue" and "headache" were not linked. \revised{We also expect that this approach can also be generalized for BioNER tasks since hyperlinks are often associated with a biomedical concept.}  

\paragraph{Training} We fine-tune a PLM with WikiHyperlinik by following the standard protocol for fine-tuning PLMs for sequence labeling tasks \citep{devlin2019bert}. Herein, for a given $N$ number of sentences, a PLM calculated the probability distribution for the $C$ classes of each token in the sentence composed of $S$ tokens. The model was trained to optimize cross-entropy (CE) loss of Eq.~\ref{equ:wiki_wikipost}, where $y_{n,s,c}$ and $\hat{y}_{n,s,c}$ indicate the label and the model's output, respectively, for the $s^{th}$ token's probability of the $n^{th}$ sentence belonging to the class $c$ respectively.

\vspace{-6mm}
\begin{equation}
\small
\label{equ:wiki_wikipost}
\mathcal{L}_{CE} = \frac{1}{N S C} \sum_{n=1}^{N} \sum_{s=1}^{S} \sum_{c=1}^{C} y_{n,s,c}\cdot\log\hat{y}_{n,s,c}
\end{equation}

\subsection{Neural Network-based Medical Jargon Extraction Model}

Our jargon prediction model consists of the following two parts: 1) additional learning feature extraction, 2) target model. This section explains the additional feature extraction at first. Then, we describes the structure of the target model.

\subsubsection{Auxiliary Feature Extraction}
The UMLS concepts incorporate important clinical domain-specific jargon, including disease, surgery, drug, etc \citep{katona2014szte, chen2018natural}. Therefore, in this study, we extracted the UMLS concepts from input sentences and then used them as features for medical jargon term extraction. We elicited UMLS concepts from input sentences with QuickUMLS, an unsupervised UMLS concept matching tool \citep{soldaini2016quickumls}. Then, we represented the positions of concepts in binary feature extraction in BIOES binary encoding. The weighting score feature was expressed by multiplying the binary encoding of a concept by the term weighting. 

In this study, we employed the widely used TF score and the masked language model (MLM) score as term weighting methods. We normalized MLM scores and TF scores to values between [0, 1] by Min-Max scaling \citep{al2006normalization}. Details on the expression of additional features are described in Appendix~\ref{apx:feature_representation}.

\paragraph{Contextualized MLM Score}
Frequency score-based methods have been widely used to extract unfamiliar or important terms, since some jargon terms can be rarely observed in the general corpus \citep{chen2018natural}. However, term frequency-based approaches do not consider contextual factors, and therefore tend to underestimate the homonym issue. Otherwise, a language model is a probability distribution over a sequence of words. We can calculate the probability of phrases or words for a given context. In particular, in MLMs, it is known that we can understand whether knowledge of a specific concept is included in PLMs by masking part of the sentence \cite{petroni2019language, kwon2019masked, zhong2021factual}.

We proposed a MLM score that is the negative likelihood of the masked tokens from a text. Eq.~\ref{equ:MLM} is the MLM score of a UMLS medical concept $c$ for a given sentence $S$. Suppose, $T_c$ is the token length of $c$ and $p$ is the starting position of $c$. Herein, we mask concept tokens $S_p...S_{p+T_c-1}$ with a special token "[MASK]" then input the masked will be $\tilde{S}$ to PLMs. As a result, we can get the probability $P(\tilde{S}_i=S_i|\tilde{S})$ of the $i^{th}$ masked token $\tilde{S}_i$ will be $S_i$ from $\tilde{S}$. Then, we calculated the MLM score of $c$ ($MLM(c,S)$) by averaging negative log likelihoods of masked tokens. Overall, when the MLM of the model is low, it means that the masked concept can be predicted easily.

\vspace{-5mm}
\begin{equation}
\small
\label{equ:MLM}
MLM(c, S) = -\frac{1}{T_c} \sum_{i=p}^{p+T_c-1} log P(\tilde{S}_i=S_i|\tilde{S})
\end{equation}

\subsubsection{Target Model}
\revised{First, an input sentence was split into subword units through a PLM tokenizer. Binary features were input to a multi-layer perceptron (MLP), mapped into the same dimension as the token embedding vector, and then added to the output of the tokenizer. The added input is input to a Wiki-trained LM then we can get hidden. 
In the Wiki's hyperlink step, the initial parameters of the PLM trained were set to the weight parameters of Wiki-trained LM.
Then, the output of the Wiki-trained LM (Hidden) was input to an MLP to create an emission score. Simultaneously, the weighted scores and the Hidden were concatenated and then input to another MLP to create the weighted emission score. We got the final emission score by adding the emission score and the weighted emission score. Then, the final emission was input to the conditional random field \citep{lafferty2001conditional} layer. Suppose $\mathbf{P}$ is the final emission, and $y$ is a sequence of output labels. Herein, we calculated the score of the sequence $y$ defined as Eq.~\ref{equ:crfs} with the transition matrix $A$. Then, we picked the optimal output sequence $\hat{y}$ from all possible sequences of labels $\mathbf{Y}$ by jointly decoding through Viterbi searching \citep{viterbi1967error}.}

\begin{equation}
    \small
    \label{equ:crfs}
    s(y)=exp\left(\sum_{i=0}^{n}\mathbf{A}_{y_iy_{i+1}}+\sum_{i=0}^{n}{\mathbf{P}_{iy_i}}\right)
\end{equation}

\begin{equation}
    \small
    \label{equ:optimal}
    \hat{y}=\underset{\tilde{y} \in \mathbf{Y}}{\argmax}~s(\tilde{y})
\end{equation}

\section{Experiment} 

\subsection{Experimental Set Up} 
%
The experiments on the WikiHyperlink span prediction were conducted on the following settings. The Wiki data consists of approximately 26M articles, which correspond to 150M sentences with 99M hyperlinks. We used BERT-cased base \citep{devlin2019bert} and RoBERTa \cite{liu2019roberta} base models. \revised{We also utilized BioClinicalBERT \citep{alsentzer2019publicly} and BioBERT \citep{lee2020biobert} which are the state-of-the-art PLMs pretrained on biomedical 
text}. We mainly set the hyper-parameters of the PLMs by following \citet{gururangan2020don}'s post-training setting. Meanwhile, each input consisted of up to 128 tokens and the learning rate was set to 5e-4. The parameters were updated every 2,048 inputs. We trained PLMs up to 50K update steps, which is slightly less than 1 epoch (about 56K; 7 days). For the remaining hyper-parameters, we used the default setting of the Transformers library \citep{wolf2020transformers}. 

To fine-tune medical jargon extract models, we used the following settings. First, the batch size and maximum epoch were set to 32 and 3, respectively, according to the PLMs' standard training setting. We set the learning rate as 5e-5 for all models. Finally, we randomly split the dataset into a 14,542 training set (80\%), a 1,817 validation set (10\%), and a 1,819 test set (10\%). Hyper-parameters and experimental models were selected with the highest performance in the validation set, and detailed results are described in Appendix~\ref{apx:detailed_experimental_setting}. 

\begin{table*}[!t]
\center
\scriptsize	
\begin{tabular}{@{}l|@{ }c@{ }c@{ }c@{ }|@{ }c@{ }c@{ }c@{ }|@{ }c@{ }c@{ }c@{ }|@{ }c@{ }c@{ }c@{ }|@{ }c@{ }c@{ }c@{ }|@{ }c@{ }c@{ }c@{}|@{ }c@{ }c@{ }c@{ }|@{ }c@{ }c@{ }c@{}}
\hline
\multicolumn{1}{@{}c|@{ }}{\multirow{3}{*}{Model}} & \multicolumn{6}{c@{ }|@{ }}{BERT}  & \multicolumn{6}{c@{ }|@{ }}{RoBERTa} & \multicolumn{6}{c@{ }|@{ }}{BioClinicalBERT} & \multicolumn{6}{c@{}}{BioBERT}\\
\cline{2-25}
\multicolumn{1}{@{}c|@{ }}{} & \multicolumn{3}{c@{ }|@{ }}{Pretrained} & \multicolumn{3}{c@{ }|@{ }}{Wiki-trained} & \multicolumn{3}{c@{ }|@{ }}{Pretrained}  & \multicolumn{3}{c@{ }|@{ }}{Wiki-trained} & \multicolumn{3}{c@{ }|@{ }}{Pretrained}  & \multicolumn{3}{c@{ }|@{ }}{Wiki-trained} & \multicolumn{3}{c@{ }|@{ }}{Pretrained}  & \multicolumn{3}{@{ }c@{}}{Wiki-trained} \\
\cline{2-25} 
\multicolumn{1}{@{}c|@{ }}{} & Prec.  & Rec.  & F1 & Prec.  & Rec.  & F1 & Prec.  & Rec. & F1 & Prec.  & Rec. & F1 & Prec.  & Rec. & F1 & Prec.  & Rec. & F1 & Prec.  & Rec. & F1 & Prec.  & Rec. & F1 \\
\hline\hline
Vanilla & 75.08 & 75.27 & 75.18 & 75.65 & 77.11 & 76.37 & 72.00 &74.94 & 73.44 & 73.06 & 76.14 & 74.57 & 77.29 & 78.06 & 77.67 & 77.39 & 79.78& 78.57 & 77.79 & 78.59& 78.19 & 78.76 &  79.40& 78.79\\
\hline
Binary & 76.44  & 77.06  & 76.75  & 76.31  & 78.51  & 77.39  & 74.00 & 75.10 & 74.54 & 74.63 & 76.31 & 75.46 & \textbf{78.47 }& 78.85 & 78.66 & \textbf{78.80}& 79.38 & 79.09 & 78.88 &78.34 & 78.61 & 78.76 &  79.40& 79.08\\
\hline
+TF  &  75.96  & 78.04  & 76.99  & 75.92  & \textbf{79.31}  & 77.58  & 73.90 & 76.50 & 75.18 & 74.29 & 76.44 & 75.35 & 78.16 & 79.60 & 78.88 &78.63&79.80& 79.21 & 78.73 & 79.51 & \textbf{79.12} & 78.69 & 80.01 & 79.35\\
+MLM & 76.05 & 78.09  & 77.06  & 75.83  & 79.16  & 77.46  & 72.91 & 76.00 & 74.42 & 74.36 & 76.45 & 75.39 & 78.15 & 79.65 & \textbf{78.89} & 78.75 &79.70&79.22& 78.68 & 79.45 & 79.06 & 78.65 & 80.04 & 79.34 \\
+TF+MLM  & \textbf{76.27} & 77.39  & 76.83  & \textbf{77.67}  & 77.26  & 77.26  & 73.06 & 76.14 & 74.57 & 74.12 & 76.676 & 75.37 & 78.16 & 79.02 & 78.59 &78.37&79.09&78.73& 78.49 &78.78 & 78.64 & 78.60 & 79.36 & 78.98 \\
\hline
Ensemble & 76.07  & \textbf{78.62} & \textbf{77.33} & 76.39 & 78.90 & \textbf{77.62} & \textbf{74.09} & \textbf{76.83} & \textbf{75.44} & \textbf{74.80} & \textbf{76.96} & \textbf{75.86} & 77.93 & \textbf{79.88} & \textbf{78.89} & 78.67 & \textbf{79.81} & \textbf{79.23} & \textbf{78.73} & \textbf{79.52} & \textbf{79.12} & \textbf{78.71} & \textbf{80.06} & \textbf{79.38}\\
\hline
\end{tabular}
\caption{The precision (Prec), recall (Rec) and F1 scores of MedJEx models.}
\label{tab:experimental_results}
\end{table*}

Moreover, a Wordfreq library \citep{robyn_speer_2018_1443582} was adopted to calculate TF of the candidate UMLS concepts. We performed Student's t-test \citep{student1908probable} to assess whether the change in performance between experimental results was statistically significant. Finally, we used the F1 score \citep{kwon2019effective} to evaluate the performance of the model.

\subsection{Experimental Results}

The PLMs can be categorized as the following two types: 1) \textbf{pretrained} models were initialized with standard pretrained models and 2) \textbf{Wiki-trained} models were initialized with the Wiki's hyperlink trained models. \textbf{Vanilla} models do not incorporate the UMLS features. The \textbf{binary} model has only the binary features. \textbf{+TF} and \textbf{+MLM} indicate that adding the TF score feature and MLM score feature, respectively. \textbf{+TF+MLM} concatenates two features as the weighted input. Finally, The \textbf{Ensemble} is a weighted voting of the predictions of four models (Binary, +TF, +MLM, +MLM+TF) designed to reflect various aspects of the features. The algorithm for Ensemble is described in Appendix~\ref{apx:ensemble}.

\subsubsection{Experimental Results on the Hyperlink Training Step} 
\label{sec:exp_result_Wiki_training}

\begin{figure}[!t]
\centering
\includegraphics[width=\linewidth]{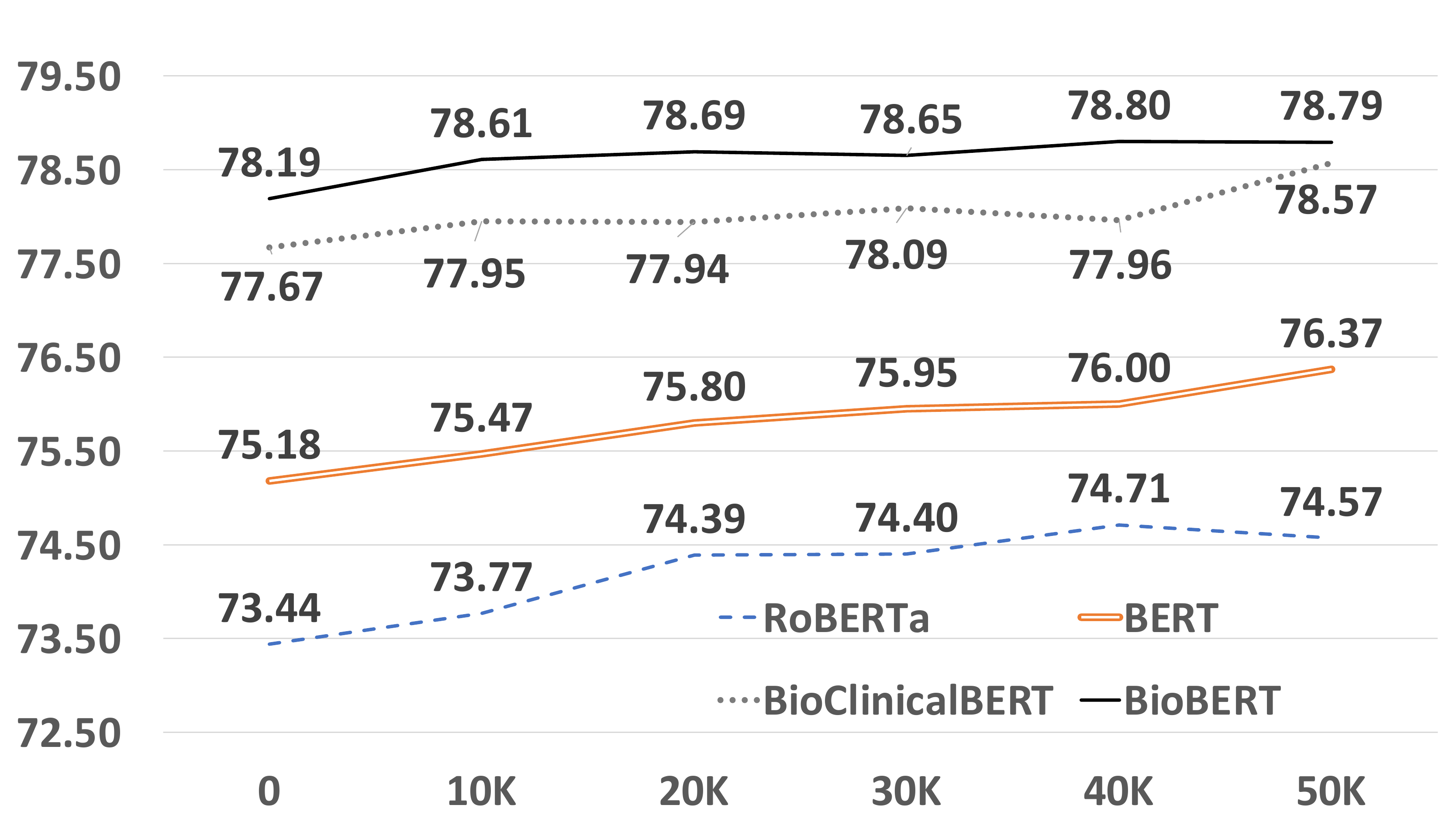}
\caption{The F1 scores of the vanilla models with the update step of the WikiHyperlink training.}
\label{fig:wiki_transfer_performance}
\end{figure}

Figure~\ref{fig:wiki_transfer_performance} is the fine-tuning performance of the vanilla models for every 10K update step on the test set. Herein, step 0 indicates the pretrained setting. \revised{The results show that the medical jargon term extraction performance tends to be improved as the update step increases. The performance was improved by 1.19\%p ($p$<1e-4) in BERT, 1.13\%p ($p$<1e-3) in RoBERTa, 0.90\%p ($p$=0.11) in BioClinicalBERT and 0.60\%p ($p$=0.07) in BioBERT, although the models were trained with less than 1 epoch of Wiki data. Considering that the pretraining corpora of all models include the English Wikipedia corpus \cite{liu2019roberta}, we can infer that the improvements are due to the hyperlink span information rather than Wiki's text. Otherwise, the performances of biomedical BERTs are marginally enhanced. We speculate that this is because the models have already been trained on the biomedical literature, so the effect of the task transfer through learning WikiHyperlink span information is relatively small. Nevertheless, these results imply that the task transfer is effective albeit Wiki data are a general corpus. Overall, we can see that the Wiki-training is beneficially transferred to medical jargon extraction models, supporting our assumption.}

\subsubsection{Impact of the Proposed Methods}
\revised{Table~\ref{tab:experimental_results} contains the experimental results for evaluating the impact of the proposed methods. Compared to the vanilla models, the Wiki-trained ensemble models outperformed by 2.44\%p in BERT ($p$<1e-11), 2.42\%p ($p$<1e-9) in RoBERTa, 1.56\%p ($p$<1e-5) in BioClinicalBERT and 1.19\%p ($p$<1e-3) in BioBERT.} We can see that the Wiki-trained models improved performance in all 24 cases compared to the pretrained models. This means that the WikiHyperlink span's information is helpfully transferred to training medical jargon. In addition, the binary models demonstrate better performance compared to the Vanilla models. Compared to Binary, TF and MLM features improve performance marginally in BERT, BioClinicalBERT and BioBERT. On the other hand, in the RoBERTa model, while the TF feature improves the performance in the pretrained model, it can be seen that the performance is slightly decreased in other cases. In addition, when both TF and MLM features are included, the performance is marginally changed compared to using each feature. The ensemble models lead to the highest performance in all cases.

\begin{table}[]
\centering
\small
\begin{tabular}{c|ccc}
\hline
          & Prec. & Rec.  & F1    \\
\hline\hline
QuickUMLS & 21.69 & 62.21 & 32.16 \\
MedCAT & 45.89 & 32.32 & 37.93 \\
\hline
\end{tabular}
\caption{The precision (Prec), recall (Rec) and F1 scores on UMLS concept extraction systems.}
\label{tab:MedJex_on_UMLS_matching_systems}
\end{table}

\subsection{Comparison with UMLS Concept Extractors}
\revised{To verify that our task is different from existing UMLS concept extraction task.} \revised{For this, we evaluated the performance of existing UMLS concept extractors, QuickUMLS and MedCAT \citep{kraljevic2019medcat}, in MedJ's test dataset. The results in Table~2 show that the performance of UMLS extractors was substantially inferior to our models. Even though QuickUMLS extracted all possible UMLS concepts, the recall score was 62.21, indicating that a substantial amount of medical jargon terms (37.79\%) in EHR notes is not included in the UMLS concepts.} \revised{To put it differently, since UMLS concept extractors mainly concentrated on specific types of medical terms, there are some representative jargon types that the UMLS concept extractors frequently fail to predict: 1) abbreviations (e.g., yo: years old, s/p: status post ...), 2) special numerical terms (e.g., 20/40: vision test results, 2-0: a heavy thread used for stitching ...). } 

\begin{table}[]
\footnotesize
\begin{tabular}{@{ }l|l|c|c@{ }}
\hline
Type        & Datasets & Pretrained & Wiki-trained  \\
\hline
\hline
\multirow{2}{*}{Disease}      & NCBI disease              & 87.92      & \textbf{89.21} \\
                              & BC5CDR disease            & 83.93      & \textbf{84.87} \\
\hline                              
\multirow{2}{*}{\begin{tabular}[c]{@{}c@{}}Drug \&\\ Chem.\end{tabular}}    & BC5CDR Chem.           & 92.07      & 91.88          \\
                              & BC4CHEMD                  & 90.06      & \textbf{90.27} \\
\hline                              
\multirow{2}{*}{\begin{tabular}[c]{@{}c@{}}Gene \&\\ Protein\end{tabular}} & BG2GM                     & 82.30      & \textbf{83.06} \\
                              & JNLPBA                    & 74.95      & \textbf{77.95} \\
\hline                              
\multirow{2}{*}{Species}      & LINNAEUS                  & 87.59      & \textbf{89.87} \\
                              & S800               & 74.95      & 74.85    \\
\hline                             
\end{tabular}
\caption{F1 score of BioBERT$_{Vanilla}$ models on Pretrained and Wiki-trained settings in BioNER datasets. Chem. indicates `chemical.'}
\label{tab:bioner_fintune}
\end{table}

\subsection{Impact of the Wiki's Hyperlink Span Training on BioNER Datasets}

\revised{We assessed the generalizability of Wiki training by conducting experiments on eight BioNER benchmarks used in \citet{lee2020biobert}'s setting \footnote{NBCI disease \citep{dogan2012improved}, BC5CDR disease \citep{wei2016assessing}, BC5CDR chemical \citep{wei2016assessing}, BC4CHEMD chemical, \citep{krallinger2015chemdner}, BC2GM \citep{smith2008overview}, JNLPBA \citep{kim2004introduction}, LINNAEUS \citep{gerner2010linnaeus}, S800 \citep{pafilis2013species}}. We evaluated the performance of the BioBERT$_{Vanilla}$ model on each data in the pretrained and Wiki-trained settings.} \revised{Table~\ref{tab:bioner_fintune} shows F1 scores on the datasets. We can see that Wiki training positively affected five datasets while it marginally impacted three datasets (S800, BC5CDR Chem. and BC4CHEMD). Especially, in JNLPBA, the performance improveed by 3\%p, and in LINNAEUS, the performance improved by 2.29\%p.} \revised{Meanwhile, since BioNER benchmarks are targeted to elicit specific medical concepts, there is some medical jargon that BioNER cannot cover, such as metric units (millliter, mg, ...) and medical techniques (flushing: to use fluid to clean out a catheter) and so on. Detailed experimental settings and results are described in Appendix~\ref{apx:bioner_finetune}.} 

\section{Discussion} 


\subsection{Feature Analysis}
\begin{figure}[!t]
\centering
\includegraphics[width=.8\linewidth]{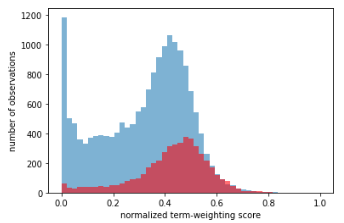}
\caption{Histograms of the MLM score feature for UMLS biomedical concepts. Red: jargon concepts; blue: non-jargon concepts}
\label{fig:overlapped_mlm}
\end{figure}

\paragraph{MLM}
Figure~\ref{fig:overlapped_mlm} represents the histograms of the biomedical concepts on MLM scores. The blue-colored histogram indicates the UMLS biomedical concepts that are not jargon, while the red-colored histogram indicates UMLS biomedical concepts that are jargon. We can notice the heavily tail of the histogram of non-jargon concepts indicating MLM scores are lower. The heavy tail includes concepts that are relatively easy to understand (e.g. shoulder, chest pain, wound management ...). These results show that the MLM score can be an appropriate feature to determine whether a concept will be the medical jargon. Additional analyses for the MLM score are in Appendix~\ref{apx:additional_analysis}.

\paragraph{TF and MLM}
We conducted a case study to analyze the impact of the TF and MLM for identifying medical jargon from EHR notes. Specifically, we calculated the TF and MLM scores of candidate UMLS concepts that had been mapped to the medical jargon in our EHR note sentences. Then, we categorized the concepts according to their scores. We used MLM of the pretrained BERT, and the concepts were categorized as a high score (> 0.5; $\uparrow$) and a low score (< 0.5; $\downarrow$). Note that a high MLM score means that the BERT failed to predict the concept. A high TF score means that the concept was frequently observed in the general corpus. The following is the combination of MLM and TF categories and notable examples.

\begin{enumerate}[leftmargin=*]
\vspace{-3mm}
    \item $\uparrow$ TF, $\uparrow$ MLM: "shock", "drainage", "tissue"
\vspace{-3mm}
    \item $\downarrow$ TF, $\uparrow$ MLM: "Vancomycin B", "Seroquel", "subdural hematoma"
\vspace{-3mm}
    \item $\uparrow$ TF, $\downarrow$ MLM: "coma", "gene", "wound"
\vspace{-3mm}
    \item $\downarrow$ TF, $\downarrow$ MLM: "pneumonia", "membrane", "viral"
\end{enumerate}

\revised{The first case is a word frequently observed in a general corpus ($\uparrow$ TF), but it is a concept that fails to predict in the BERT ($\uparrow$ MLM). This concept includes rare senses used in medical contexts. For example, "drainage" is used as a synonym for sewer in the general context, while in the medical domain it may mean "extra liquid that is removed from the body." The second case is the most unfamiliar words. The concepts are composed of multiple tokens and medical entities such as disease or drug names.
The third case consists of relatively easy-to-understand concepts.
The fourth case contains relatively short medical jargon composed of 1 to 2 tokens. We can infer that MLM and TF cannot only be complementary but also can be used together to help solve the challenging homonym issue. }

\subsection{Error Analysis}

We manually examined the outputs. The most common type of false negatives errors was abbreviations, such as "ENT" for "ear, nose, and throat" and "or" for "operating room", and "p.o." for "per os". Another type of error was signs with special meanings such as "q.6 h" for "per every 6 hours", "x2" for "two times", and "3+" for "very strong" fall into this type. Other notable errors were epoymous person name-based medical concepts (e.g., "Azzopardi effect") and device names (e.g., "BiPap"). 
MedJEx failed to detect the aforementioned types of medical jargon due to the data sparsity challenge. 

\subsection{Prospective Downstream Applications} This new method for medical jargon extraction has divers potential applications. It could be a preprocessing part of BioNLP pipelines and used for downstream medical AI application systems. For example, it could be adapted to medical concept linking systems such as NoteAid \citep{polepalli2013improving}. In addition, a chatbot-based self-diagnosis system \citep{you2020self} could use our approach for the explanation of medical jargons to avoid generating jargons. 

\subsection{Merits} \revised{Prospective downstream applications can promote effective communication between clinicians and their patients by increasing patients' EHR comprehension ability. This, in turn, can help the patients in self-management of their illness \citep{adams2010improving}. Effective communication is also beneficial for preventing physicians' burnout \citep{aaronson2019training}. Thus, we can expect this new task will contribute not only to improve the patients' outcomes.} 

\subsection{Limitations} 
\revised{This task defined medical jargon at a single difficulty-level, disregarding diverse educational levels of users. In particular, setting the difficulty of each medical jargon term will help this task contribute to improving the performance of machines as well as patients, and further educate and support clinicians. Moreover, we did not analyze the jargon types such as acronyms but merely identified the presence of medical jargon, which can limit further analyses.}


\section{Conclusion} 
\label{sec:conclusion_and_future_work}
We introduce a novel NLP tasked named MedJEx and present an expert-curated MedJ dataset for the task. 
We propose two innovative methods: 1) Pretraining Wiki's hyperlink span, and 2) Contextualized MLM score feature for extracting medical jargon from EHR notes. The experimental results show that the Wiki's hyperlink span can be effectively transferred to the medical jargon extraction model, leading to a significant performance improvement. Wiki's hyperlink span training also beneficial in six out of eight BioNER benchmarks. Finally, in a qualitative evaluation, the MLM score feature complements the TF feature to identify common terms (or terms with high TFs) used in the clinical domain (homonyms). 



\section*{Ethical Consideration}
\revised{In this study, we legitimately obtained a licensed access to the University of Pittsburgh Medical Center EHR repository, and all EHR notes used were fully de-identified.The experiments described in Appendix~\ref{apx:annotation_reliability} and \ref{apx:user_evaluation} were performed in accordance with the recommendations laid out in the World Medical Association Declaration of Helsinki. The study protocol was approved by the institutional review boards of a medical school in the US.}

\revised{In addition, our model used BERT and its families, so it over-relies on a contextual embedding feature that can cause mis-classification. Specifically, even with the same terminology, the prediction of a model may be different depending on the context.}

\revised{There are some obstacles to re-implement our work. First of all, training Wiki's hyperlink takes a week to train with an RTX8000 GPU and computing cost can be problematic. Therefore, we will make our Wiki-trained models publicly available. Second, we used UMLS concept extractors (QuickUMLS and MedCAT) which require a license from the National Library of Medicine.}

\section*{Acknowledgement}
Research reported in this study was in part supported by the National Institutions of Health R01DA045816 and R01MH125027. The content is solely the responsibility of the authors and does not necessarily represent the official views of National Institutes of Health.

\bibliography{anthology,custom}
\bibliographystyle{main}


\setcounter{figure}{0} 
\setcounter{table}{0} 
\newpage
\appendix
\onecolumn
\section*{Appendices}

\section{Data Annotation}
\label{apx:data_construction}

\subsection{Evaluation of the Annotation}
\label{apx:annotation_reliability}
\revised{To evaluate the annotators' reliability in identifying jargon, an observational study was performed to assess the agreement of the dataset annotators with each other and with laypeople. \textbf{Note that, this work is a part of unpublished manuscript.}}


\subsubsection{Data Collection and Setting}
For evaluation, twenty sentences were randomly selected from deidentified inpatient EHR notes from the University of Pittsburgh Medical Center EHR repository.
 Sentences that consisted only of administrative data, sentences whose length was 
less than ten words, and sentences substantially indistinguishable from another sentence were filtered out. 

Note that, the annotators had never seen the sampled sentences. The twenty sentences were made up of 904 words in total. Common words were discarded so as not to inflate the calculated agreement. These consisted of all pronouns, conjunctions, prepositions, numerals, articles, contractions, months, punctuation, and the most common 25 verbs, nouns, adverbs, and adjectives. Terms occurring more than one time in a sentence were counted only once. Furthermore, to ameliorate double-counting issue, multi-word terms were counted as single terms. Multi-word terms were determined by two members of the research team by consensus. In this work, multi-word terms were defined as adjacent words that represented a distinct medical entity (examples: “PR interval”, “internal capsule”, “acute intermittent porphyria”), were commonly used together (examples: “hemodynamically stable”, “status post”, “past medical history”) and terms that were modified by a minor word (examples: “trace perihepatic fluid”, “mild mitral regurgitation”, “rare positive cells”, “deep pelvis”). After applying these rules, 325 candidate medical jargon terms were utilized. The laypeople consisted of 270 individuals recruited from Amazon Mechanical Turk (MTurk) \citep{aguinis2021mturk}.

\subsubsection{Annotation Reliability}
The results showed that there was good agreement among annotators (Fleiss’ kappa = 0.781). The annotators had high sensitivity (91.7\%) and specificity (88.2\%) in identifying jargon terms as determined by the laypeople (the gold standard).

\section{Details on Feature Representations}
\label{apx:feature_representation}

\begin{figure*}[!ht]
\centering
\begin{subfigure}[b]{0.49\textwidth}
\includegraphics[width=\linewidth]{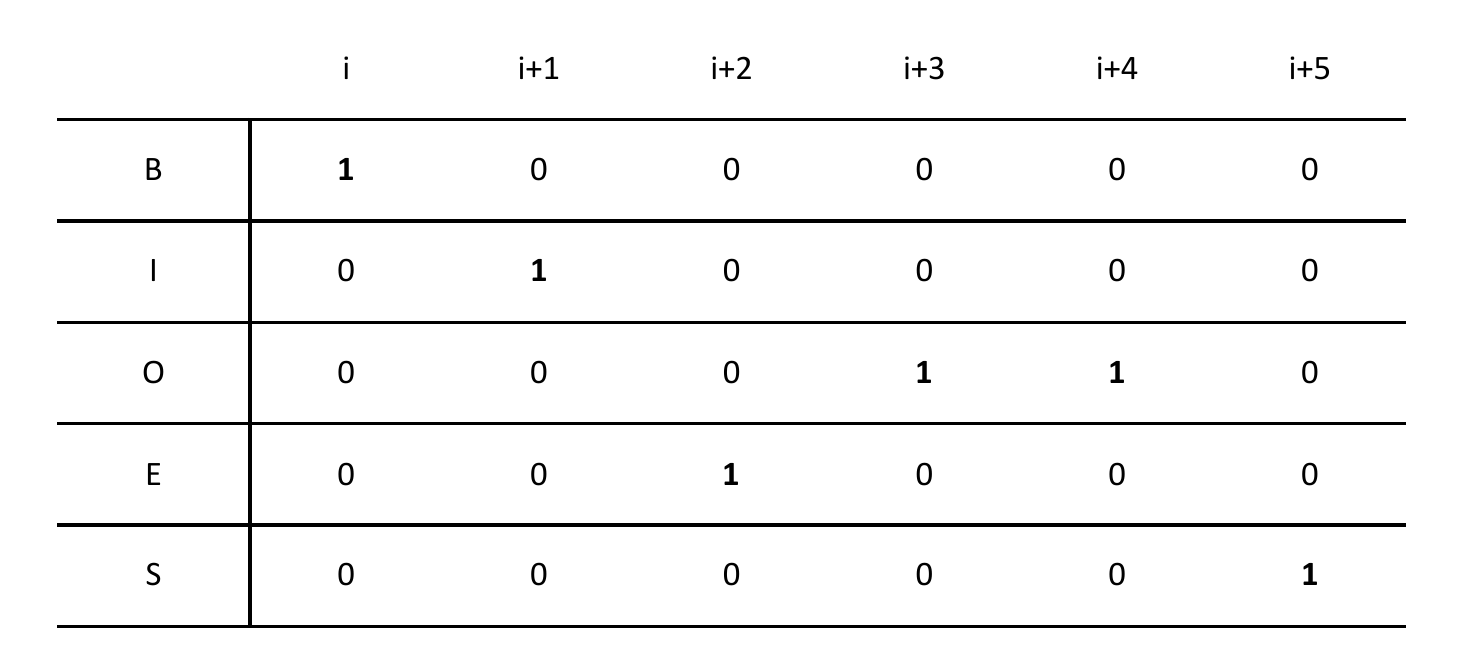}
\caption{Binary feature}
\label{fig:binary_feature}
\end{subfigure}
\begin{subfigure}[b]{0.49\textwidth}
\includegraphics[width=\linewidth]{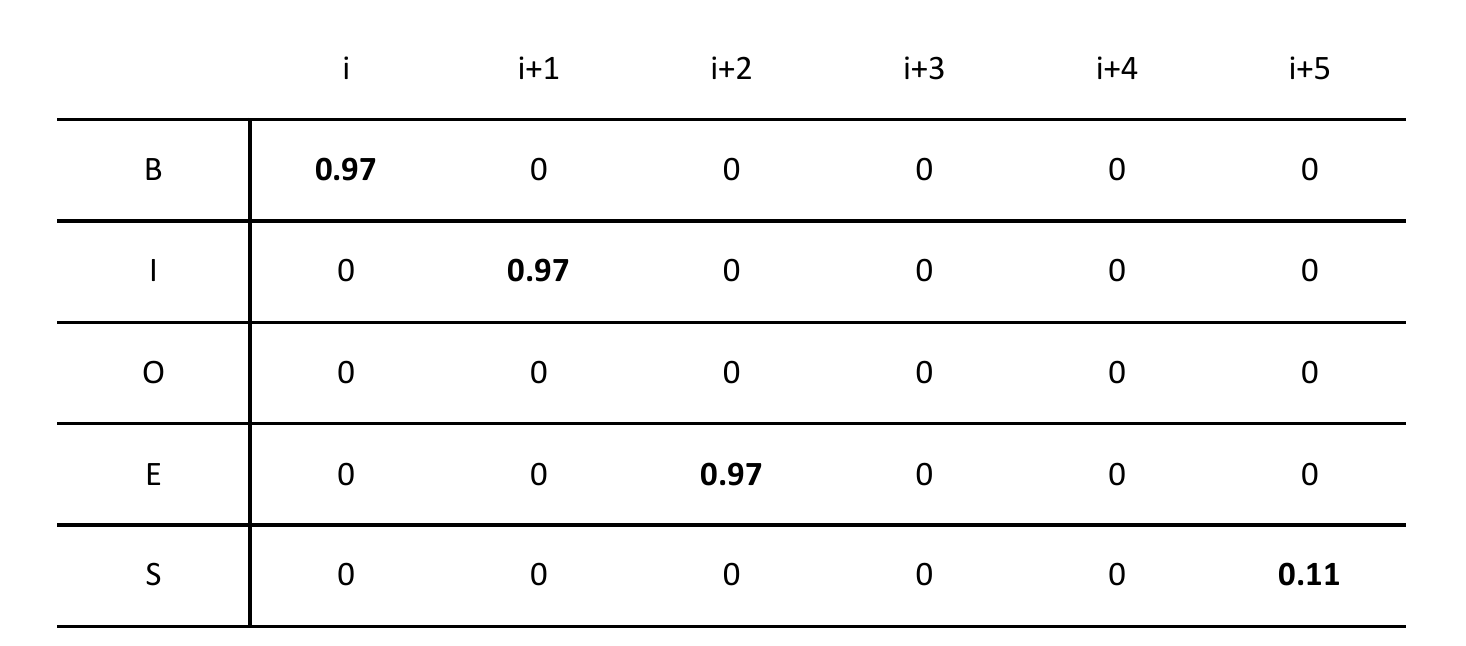}
\caption{Weighted score feature}
\label{fig:weighted_feature}
\end{subfigure}
\caption{Examples of a binary feature and a weighted score feature.}
\end{figure*}

\revised{This section explains details on the binary and weighted score features with an example. Suppose a concept $c_1$ starts with the $i^{th}$ token and the length is $3$, and another concept $c_2$ starts with the $i+5^{th}$ token and the length is $1$.
In this case, binary encoding features can be expressed as a 5-dimensional vector ("B", "I", "O", "E", "S") as shown in Figure~\ref{fig:binary_feature}. Since $c_1$ starts at $i$ and ends at $i+2$, the $i^{th}$ "B" dimension and $i+2^{th}$ "E" dimension are set to 1. In addition, because the $i+1^{th}$ token is the inner of the $c_1$, the $i+1^{th}$ `I' dimension will be 1. Then, $c_2$ starts and ends at $i+5$.}
On the other hand, further assume that the term weighting score of c1 and c2 are 0.97  and 0.11

\section{Details of the Experimental Setting}
\label{apx:detailed_experimental_setting}
In the experiments, we set the models’ hyper-parameters with the performance on the validation set via the grid-search.  Once, determining hyper-parameters in vanilla models, the same values were used in the other models.  We  trained WikiHyperlink for 50K update steps.  Herein, the  number of parameters of all experimental models are about 108M. In all experiments, the random seed was set to 0. All experiments were conducted in the Centos Linux 7 environment using one RTX-8000 GPU, Intel Xeon E5-2620 CPU, and 64GB RAM.

\subsection{Hyper-parameter Setting}

In the case of MLP, all hidden sizes were set equal to the default hidden size of PLM. Also, the activation function used a hyperbolic tangent function by following \citet{gururangan2020don}'s setting.

\begin{table}[!ht]
\centering

\begin{tabular}{c|c|c|c|c}
\hline
  Learning Rate  & BERT  & RoBERTa & BioClinicalBERT & BioBERT \\
\hline\hline
6e-5 & 67.46 & 62.80 & 67.07 & 70.48\\
\hline
1e-5 & 71.02 & 66.91 & 73.06 & 73.97\\
\hline
5e-5 & \textbf{75.92} & \textbf{73.64} & \textbf{78.53} & \textbf{78.51}\\
\hline
\end{tabular}
\caption{The F1 scores  of the finetuned vanilla models for each learning rate.}
\label{tab:validation_finetuning_learning_rate}
\end{table}

In the fine-tuning on the task, we choose the best learning rate of the vanilla models on the validation set among the following set of the candidate learning rates \{5e-6, 1e-5, 5e-5\}. Overall, the results in Table~\ref{tab:validation_finetuning_learning_rate} show that we could achieve the best performances on validation set when the learning rate was set 5e-5.

\subsection{Model Selection}

\begin{table}[!ht]
\centering

\begin{tabular}{c|c|c|c}
\hline
  BioClinicalBERT & BioBERT & BioMedRoBERTa  & BioClinicalRoBERTa  \\
\hline\hline
\textbf{78.53} & 78.51 & 73.89 & 76.27 \\
\hline
\end{tabular}
\caption{The F1 scores of the fine-tuned pretrained biomedical PLMs on vanilla setting.}
\label{tab:model_selection}
\end{table}

\revised{In this work, we utilize contextualized PLMs to make jargon prediction models. For this, we use two representative PLMs: BERT \citep{devlin2019bert} and RoBERTa \citep{liu2019roberta}. In addition, there are several recent state-of-the-art models pretrained in biomedical domains, recently. To be specific, BioBERT additionally trained BERT with biomedical text corpora  \citep{lee2020biobert}. BioClinicalBERT \citep{alsentzer2019publicly} further trained BioBERT with clinical notes from MIMIC-III \citep{johnson2016mimic}. In addition, there are some studies, such as BioMedRoBERTa \citep{gururangan2020don} or BioClinicalRoBERTa \citep{lewis2020pretrained} that suggested training the RoBERTa model with biomedical text corpora  or clinical notes. On the other hand, \citet{michalopoulos2021umlsbert} proposed UmlsBERT that integrates UMLS semantic type embedding as an additional input feature during the pretraining step. UmlsBERT is similar to our suggestion in that it uses UMLS concept as an embedding feature. However, our method is slightly different in that it uses span information instead of the UMLS semantic type. Moreover, we show that the performance can be improved using the UMLS features only in fine-tuning.}

In this paper, we selected two biomedical PLMs by comparing the performances of state-of-the-art biomedical PLMs in the vanilla models on the validation set. Table~\ref{tab:model_selection} presents the experimental comparison among the four representative biomedical PLMs: BioClinicalBERT \citep{alsentzer-etal-2019-publicly}, BioBERT \citep{lee2020biobert}, BioMedRoBERTa and BioClinicalRoBERTa. The results show that BioBERT and BioClinicalBERT showed no differences ($p>0.05$) but the other RoBERTa-based models presented inferior scores. Therefore, we choose the BERT-basedbiomedical PLMs for further experiments.




\newpage
\subsection{Experimental Results of UMLS concept extractors}

\begin{table}[!h]
\centering
\begin{tabular}{c|ccc}
\hline
          & Prec. & Rec.  & F1    \\
\hline\hline 
QuickUMLS & 21.10 & \textbf{60.74} & 31.32 \\
MedCAT    & \textbf{45.89} & 32.32 & \textbf{37.93} \\
\hline
\end{tabular}
\caption{The precision (Prec), recall (Rec) and F1 scores on UMLS concept extraction systems.}
\label{tab:MedJex_on_UMLS_matching_systems}
\end{table}

\begin{table}[!h]
\centering
\begin{tabular}{c|c|ccc}
\hline
    Setting & Concept Extractor & Prec. & Rec.  & F1    \\
\hline\hline
BERT$_{Binary}$ & QuickUMLS & \textbf{75.51} & \textbf{77.37} & \textbf{76.43} \\
BERT$_{Binary}$ & MedCAT    & 74.90 & 77.00 & 75.93 \\
\hline
\end{tabular}
\caption{The precision (Prec), recall (Rec) and F1 scores on BERT with a binary setting on the different concept extractors.}
\label{tab:BERT_Binary_on_UMLS_matching_systems}
\end{table}


\revised{Since experimental models rely on a UMLS concept extractor, it is also important to choose appropriate UMLS concept extractors. There are several concept extractors that have been introduced including MetaMap \citep{demner2017metamap}, QuickUMLS \citep{soldaini2016quickumls}, cTAKES \citep{saputra2018keyphrases}, and MedCAT \citep{kraljevic2021multi}. We compared two extractors, QuickUMLS and MedCAT, which are state-of-the-art concept extractors. Table~\ref{tab:MedJex_on_UMLS_matching_systems} presents the performance of the concept extractors. Herein, we can see that MedCAT achieved better performance in terms of precision and F1 but QuickUMLS had better recall performance. We preferred higher recall, since a concept extractor was used for candidate concept extraction. Indeed, the performances on the BERT with the binary setting in Table~\ref{tab:BERT_Binary_on_UMLS_matching_systems} demonstrates that using QuickUMLS led to higher performance than that of using MedCAT.}

\subsection{Experimental Results on Tagging Schemes}
\begin{table}[!h]
    \centering
    \begin{tabular}{c|c}
       \hline
       Tagging scheme & F1\\
       \hline\hline
       BIO  & 74.25 \\
       \hline
       BIOES & \textbf{75.92}\\
    \hline
    \end{tabular}
    \caption{Experimental comparison on BIO and BIOES tagging schemes.}
    \label{tab:tagging_scheme}
\end{table}

\revised{Finally, to select a sequence labeling tagging scheme, we compared two representative tagging schemes: Begin, Inside and Outside (BIO) and Begin, Inside, Outside, End, and Singleton (BIOES) \citep{yang2018design}. Table~\ref{tab:tagging_scheme} presents the experimental results on the validation set of the BERT's vanilla setting. The results show that the validation performance with the BIO scheme is lower than that of the BIOES scheme.}

\newpage
\section{The Impact on the Understandability to Patients}
\label{apx:user_evaluation}
This section introduces an experiment to verify that providing medical jargon and the corresponding lay definitions can be beneficial to the comprehension of patients. Note that all experimental settings and results are part of \citet{lalor2021evaluating}'s work.

\subsection{Experimental Setting}
The authors recruited 174 patients from a community hospital in the USA. Herein, the participants took a web-based EHR comprehension test and the participants were randomly assigned to a control (n=85) group or intervention (n=89) group to take the test without or with the support of the medical jargons identification and the corresponding lay definitions, respectively. In addition, 200 participants from MTurk were engaged to take the test (100 participants were assigned to a control group and the other 100 were allocated to an intervention group).

\subsubsection{EHR Comprehension Test}

\begin{figure}[h]
    \centering
    \includegraphics[width=.8\linewidth]{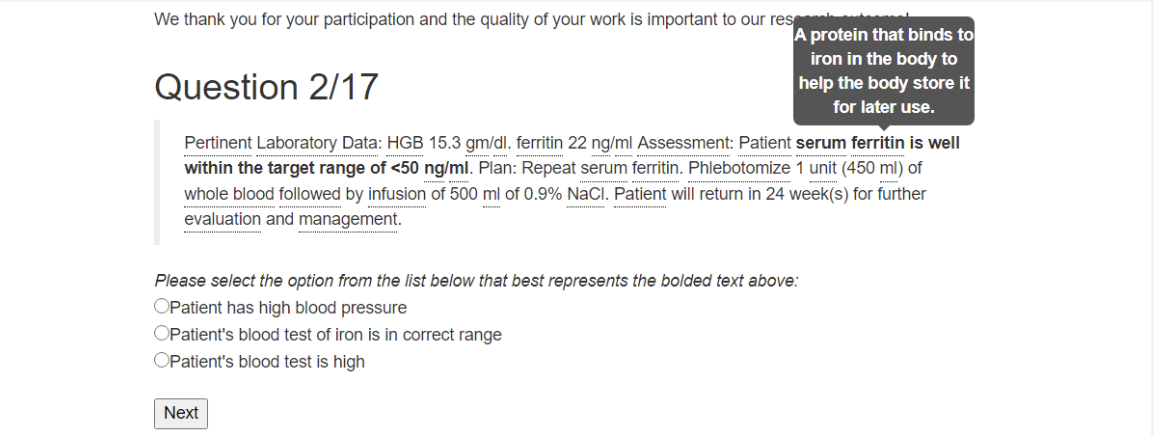}
    \caption{An example question of the EHR comprehension test. Herein, you can see that identifying medical jargon ``ferritin'' and providing its definition can be helpful to understand that the bold text describes a blood iron test.}
    \label{fig:comprehENotes}
\end{figure}

To assess a user's comprehension of EHR notes, we conducted the EHR comprehension test. Table~\ref{fig:comprehENotes} is an example of the EHR comprehension test. This test consists of 14 paragraphs extracted from de-identified EHR notes and relevant multiple-choice questions curated by physicians. In previous work, it has been verified that the EHR comprehension test reflects the participant's education level and understandability of the medical literature. In this experiment, we provided definitions for medical jargon only to the intervention group.  

\subsection{Experimental Results}
\begin{table}[!h]
    \centering
    \begin{tabular}{c|c|c}
        \hline
         Source & Condition control & Intervention \\
        \hline\hline 
         MTurk & 0.756$\pm$0.246 & 0.830$\pm$0.201 \\
         \hline
         Local hospital & 0.646$\pm$0.179 & 0.727$\pm$0.191 \\
         \hline
    \end{tabular}
    \caption{Experimental evaluation on patients' understandability. Herein, the values in the table indicate the average score and standard deviation of each group on the EHR comprehension test.}
    \label{tab:understandability_evalutaion}
\end{table}
Table~\ref{tab:understandability_evalutaion} presents the experimental results of the evaluation of the customers' understandability. The results of ANOVA \citep{cuevas2004anova} show that providing medical jargon and the corresponding lay definitions significantly enhances the patients' comprehension of the EHR notes in both groups ($p<0.01$). 

\newpage
\section{Algorithm for the Ensemble Model}
\label{apx:ensemble}

\begin{algorithm}
    \label{algo:man_itg}
    \begin{algorithmic}[1]
        \caption{Pseudo code for the weighted voting ensemble prediction}
        \Require Validation Set (\textit{V}), Test Set (\textit{T}), Trained Models (\textit{M})
        
        \State $S \gets \varnothing$ 
        \For{$M_i$ in $M$}
            \State $L \gets \varnothing$
            \For{$V_j$ in $V$}
                \State $\textbf{X}_{V_j}, \textbf{y}_{V_j} \gets V_j$
                \State $L \gets L ||  M_i(\textbf{X}_{V_j})$
            \EndFor
            \State $S \gets S || F1(L, V)$
        \EndFor
        
        \State $O \gets \varnothing$
        \For{$T_j$ in $T$}
            \State $L \gets \varnothing$
            \For{$M_i$ in $M$}
                \State $\textbf{X}_{T_j}, \textbf{y}_{T_j} \gets T_j$
                \State $L \gets L + S_i \times M_i(\textbf{X}_{T_j})$
            \EndFor
            
            \For{$L_k$ in $L$}
                \State $O \gets O|| \argmax$ $L_k$
            \EndFor
        \EndFor
        
        \State \textbf{Return} $O$
        
    \end{algorithmic}
\end{algorithm}

We first evaluated the performance on the validation set of each model and then set this as the weight of each model (line 2 to 11). In the line 1, we set of models' F1 scores $S$ as $\varnothing$. Herein, in line 7, we got the $j^{th}$ sequence of optimal predicted labels of a model ($M_i(\textbf{X}_{V_j})$) and appended it to the list of predictions $L$. Then, we calculated $F1$ score of each model ($M_i$) for the given validation set $V$ in line 8. 

In the test set, weighted voting was performed based on the scores of the models (line 11 to 20). In line 15, we got the result of multiplying the model's score by the predicted sequence of labels ($S_i \times M_i(\textbf{X}_{T_j})$) for a test input ($\textbf{X}_{T_j}$). After that, the label with the highest weighted score was selected from each token in line 18. Finally, the selected labels $O$ are returned in line 21.

\newpage

\newpage
\section{Impact of the Proposed Methods on BioNER Benchmarks}
\label{apx:bioner_finetune}

\begin{table}[!ht]
\centering
\begin{tabular}{c|l|l|ccccc}
\hline
\multicolumn{1}{c|}{Setting}  & \multicolumn{1}{c|}{Entity Type}  & Datasets  & Vanilla  & Binary  & TF  & MLM  & TF+MLM  \\
\hline
\hline
\multirow{8}{*}{Pretrained}  & \multirow{2}{*}{Disease}  & NCBI disease  & 87.92  & 87.62  & 88.31  & 88.24  & 87.53  \\
  &  & BC5CDR disease  & 83.93 & 84.10  & 84.62  & 84.74  & 83.96  \\
\cline{2-8}
  & \multirow{2}{*}{Drug/chem}  & BC5CDR chemical & 92.07 & 91.58  & 92.08  & 92.09  & 91.64  \\

  &  & BC4CHEMD  & 90.06  & 90.30  & 90.04  & 89.97  & 90.14  \\
\cline{2-8}
  & \multirow{2}{*}{Gene/protein} & BG2GM  & 82.30& 82.87  & 82.83  & 82.81  & 82.67  \\
  &  & JNLPBA  & 74.95 & 78.04  & 77.51  & 77.41  & 77.90  \\
\cline{2-8}
  & \multirow{2}{*}{Species}  & LINNAEUS  & 87.59 & 87.86  & 85.69  & 85.81  & 85.92  \\
  &  & Species-800  & 74.95 & 75.11  & 76.88  & 77.12  & 76.67  \\
\hline
\multirow{8}{*}{Wiki-trained} & \multirow{2}{*}{Disease}  & NCBI disease  & \textbf{89.21} & 85.86  & 87.90  & 87.85  & \textbf{88.24} \\
  &  & BC5CDR disease  & \textbf{84.87} & \textbf{85.47} & \textbf{85.23} & \textbf{85.35} & \textbf{85.62} \\
\cline{2-8}
  & \multirow{2}{*}{Drug/chem}  & BC5CDR chemical & 91.88  & \textbf{92.17} & \textbf{92.09} & \textbf{92.13} & \textbf{91.84} \\
  &  & BC4CHEMD  & \textbf{90.27} & 90.27  & \textbf{90.21} & \textbf{90.13} & \textbf{90.50} \\
\cline{2-8}
  & \multirow{2}{*}{Gene/protein} & BG2GM  & \textbf{83.06} & \textbf{83.23} & \textbf{83.07} & \textbf{83.26} & \textbf{83.17} \\
  &  & JNLPBA  & \textbf{77.95} & \textbf{78.34} & \textbf{77.69} & \textbf{78.33} & \textbf{77.70} \\
\cline{2-8}
  & \multirow{2}{*}{Species}  & LINNAEUS  & \textbf{89.87} & \textbf{88.67} & \textbf{89.10} & \textbf{88.83} & 85.85  \\
  &  & Species-800  & 74.85  & \textbf{75.76} & 75.97  & 76.12  & 75.10  \\
 \hline
\end{tabular}

\caption{Experimental results on proposed methods on BioNER benchmarks. Herein, the values are presented in \textbf{bold} if performance is improved in Wiki\_setting. }
\label{tab:proposed_methods_on_BioNER_benchmarks}
\end{table}

\revised{We examined the impact of our suggestions on BioNER benchmarks. Herein, we mainly compare the impact of WikiHyperlink span training method. Furthermore, we kept the hyper-parameters of other experiments and we used the F1 score as the main evaluation criterion.}

\revised{Table~\ref{tab:proposed_methods_on_BioNER_benchmarks} presents the experimental results. We can see that the performances were enhanced in almost all cases. Among 40 experimental settings performances were improved on 30 settings. In addition, when we conducted a t-test on the performance of the experimental settings, we found that the mean and standard deviation of the settings were significantly different ($p < 0.005$). However, in Species-800 and NCBI disease, overall performance marginally decreased after Wiki\_training application, and it can be found that the performance marginally changed in the BC4CHEMED data. }

\revised{On the other hand, additional features did not the affect performance improvement. 
This is due to the structure of our model. Note that the candidate medical terms are extracted by QuickUMLS and encoded as a binary form (see Appendix~\ref{apx:feature_representation}). 
This can be advantageous in the MedJ task that extracts comprehensive medical terms. However, the BioNER task aims to extract entities of specific semantic types. 
Therefore, our approach can confuse the NER models. For instance, ``Von Willebrand's factor deficiency'' is a syndrome name and part of it, ``Von Willebrand's factor'', is a protein name. In our setting, ``Von Willebrand's factor deficiency'' is input as a medical jargon. However, if a task is gene/protein name extraction, the input signal can mislead the model.} \revised{To ameliorate this issue, we can utilize semantic type information from biomedical concept extractors. Specifically, we can use semantic type embedding \citep{michalopoulos2021umlsbert} or a semantic-type span feature \citep{kwon2019effective} as an additional input.}

\newpage
\section{Additional Analysis of the MLM Score Feature}
\label{apx:additional_analysis}

\begin{figure*}[!ht]
\centering
\begin{subfigure}[b]{0.49\textwidth}
\includegraphics[width=\linewidth]{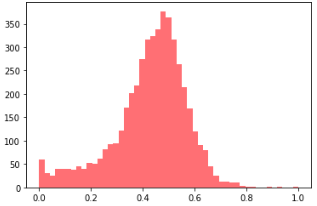}
\caption{MLM scores of UMLS concepts which are jargon}
\label{fig:jargon}
\end{subfigure}
\hfill
\begin{subfigure}[b]{0.49\textwidth}
\includegraphics[width=\linewidth]{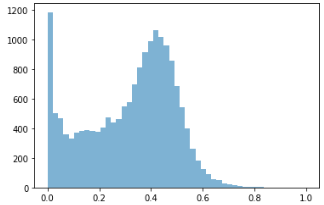}
\caption{MLM scores of UMLS concepts which are not jargon}
\label{fig:not_jargon}
\end{subfigure}
\caption{Histograms of the MLM score feature. The x-axis is the normalized MLM score, and the y-axis is the number of observations.}
\label{fig:MLM_score_features}
\end{figure*}

In this section, we show that MLM features can be valid for extracted UMLS biomedical concepts. Figure~\ref{fig:MLM_score_features} is a histogram of biomedical concepts for MLM scores. In this case, Figure~\ref{fig:jargon} is a histogram for medical jargons, and Figure~\ref{fig:not_jargon} is a histogram for non-jargons. As a result of the experiment, we confirmed that, in the case of non-medical jargon, the histogram showed a heavily tailed distribution in the section with the low MLM score. On the other hand, medical jargon was observed relatively infrequently at low MLM scores.

The mean and standard deviation of the jargon's MLM score were $0.43\pm0.14$, and the mean and variance of non-jargon concepts were $0.32\pm0.17$. As a result of performing the statistical test, we can see that the two distributions were significantly different ($p < 0.01$).

\end{document}